\documentclass{article}

\usepackage[preprint]{neurips_2020}

\newcommand{\visBasePath}[0]{vis/}
\usepackage[utf8]{inputenc} 
\usepackage[T1]{fontenc}    
\usepackage[english]{babel}
\usepackage{amsfonts}       
\usepackage{amsmath}
\usepackage{graphicx} 
\usepackage{xspace} 
\usepackage{todonotes}
\usepackage{subfig} 
\usepackage{hyperref}       
\usepackage{url}            
\usepackage{booktabs}       
\usepackage{nicefrac}       
\usepackage{microtype}      
\usepackage[linesnumbered,ruled]{algorithm2e}

\newboolean{vc_is_included}
\gdef\GITAbrHash{bb6d439}%
\gdef\VCDateTEX{2020/11/25}%
}{
	\setboolean{vc_is_included}{false}
	
	\gdef\GITAbrHash{}

	\gdef\VCDateTEX{}

}

\usepackage{amssymb}

\newcommand{\articleTitle}[0]{DeepGG: a Deep Graph Generator}

\newcommand{\articleEmail}[0]{julian.stier@uni-passau.de}

\title{\articleTitle}

\author{
	Julian Stier, Michael Granitzer\\
	University of Passau\\
	Innstraße 41, 94032 Passau, Germany\\
	\texttt{\articleEmail}
	\ifthenelse{\boolean{vc_is_included}}{\tiny\protect\\\VCDateTEX~$\sim$~\GITAbrHash}{~}
}

\begin{document}

\maketitle

\begin{abstract}
Learning distributions of graphs can be used for automatic drug discovery, molecular design, complex network analysis, and much more.
We present an improved framework for learning generative models of graphs based on the idea of deep state machines.
To learn state transition decisions we use a set of graph and node embedding techniques as memory of the state machine.

Our analysis is based on learning the distribution of random graph generators for which we provide statistical tests to determine which properties can be learned and how well the original distribution of graphs is represented.
We show that the design of the state machine favors specific distributions.
Models of graphs of size up to 150 vertices are learned.
Code and parameters are publicly available to reproduce our results.

\end{abstract}

\maketitle

\section{Introduction}\label{sec:introduction}
We aim to learn generative models of graphs from an empirical set for which we assume a characteristic underlying distribution.
These desirable models find applications in the discovery of chemical compounds such as drugs and various other application areas for which the discovery is based on profound expert knowledge, empirical testing and complex underlying rules of how the graphs emerge.

We present a generalized generative model for graphs based on DGMG \cite{li2018learning} trained on sets of representative graphs and evaluate it on existing probabilistic random graph generators with various graph sizes of up to several hundred vertices.
For this, we represent graphs in construction sequences and discuss three variants to obtain such sequences.
Construction sequences are possible representations of graphs like the commonly used adjacency matrix of a graph and explained in \autoref{sec:construction-sequences}.
Equivalent or similar representations are used by various auto-regressive models as mentioned in \autoref{sec:related-work}.
Our \textbf{contributions} comprise
\begin{enumerate}
	\item a generative model for graphs learned from graph construction sequences,
	\item an introduction to and an analysis of three variants of graph construction sequences,
	\item an analysis of generated graphs obtained through examples from graph distributions of the probabilistic models of Erdős-Rényi, Barabasi-Albert and Watts-Strogatz,
	\item and accessible source code and reproducible evaluations \footnote{Our source code is published at \url{https://github.com/innvariant/deepgg}}.
\end{enumerate}

\section{Related Work}\label{sec:related-work}
For understanding our work and its embedding in its research field, we differentiate between two families of random models of graphs:
imprecisely, we call them \textit{probabilistic} models of graphs and \textit{generative} models of graphs although usually the latter are subsets of the former and both terms refer to statistical and thus mathematical models.
Here, we consider models as \textit{probabilistic} if they consist of an algorithmic description and if their draws of random samples are based on simple distributions.
Examples for probabilistic models of graphs are the Erdős-Rényi model \cite{erdos1959random} for random graphs, a special case of exponential random graph models \cite{wasserman1996logit}, the Watts-Strogatz model for small-world networks \cite{watts1998collective}, the Kleinberg model \cite{kleinberg2000navigation}, the Barabasi-Albert model for scale-free networks \cite{albert2002statistical}
Probabilistic models of graphs provide a distribution of graphs with a usually small set of parameters controlling a core principle of the model.
A major intend to study those models are statistical analyses in network science.

We consider models as \textit{generative} models of graphs if they are intended to be able to learn characteristic distributions of graphs based on an observed set of graphs from an underlying unknown distribution of graphs.
These generative models are usually of a magnitude higher in terms of their number of parameters.
While Liao et al. \cite{liao2019efficient} differentiate between auto-regressive and non-autoregressive models, we currently prefer to categorize generative models of graphs further into popular types of deep learning models which we identified as \textit{Recurrent Neural Networks} (RNN), \textit{Variational Auto-Encoder} (VAE), \textit{Generative Adversarial Networks} (GAN), \textit{Policy Networks} (PI) or a mixture of them.
Furthermore, it is important to know of techniques such as Graph Neural Networks \cite{gori2005new}, Graph Convolution \cite{kipf2016semi} and the Message Passing Framework \cite{gilmer2017neural}, in general, as they provide embedding techniques which are now widely used among above generative models.

An early work which provides a model that ``can generate graphs that obey many of the patterns found in real graphs'' is based on kronecker multiplication \cite{leskovec2007scalable} and called \textit{KronFit}.
However, the model is fitted towards a single graph with maximum likelihood.
To our understanding the original intend of it was not to learn the hidden distribution of graphs but the exemplary graph itself and despite it can ``sample of the real graph'' by ``using a smaller exponent in the Kronecker exponentiation'' it can be assumed that the distribution is close to this particular single objective graph and not an underlying distribution of graphs from which the graph might have been sampled \cite{leskovec2010kronecker}.

The most important work related to our model and analysis are Learning Deep Generative Models of Graphs (DGMG) \cite{li2018learning}, Graph Recurrent Neural Networks (GraphRNN) \cite{you2018graphrnn} and Efficient Graph Generation with Graph Recurrent Attention Networks (GRANs) \cite{liao2019efficient}.
DGMG is the model framework we extended, generalized and provided new experiments for and which computational issues Li et al. have again tackled with the successful GRAN in \cite{liao2019efficient}.
GraphRNN \cite{you2018graphrnn} is a highly successful auto-regressive model and was experimentally compared on three types of datasets called ``grid dataset'', ``community dataset'' and ``ego dataset''.
The model captures a graph distribution in ``an autoregressive (recurrent) manner as a sequence of additions of new nodes and edges''.
Liao et al. identify some major issues of GraphRNN and explain that ``handling permutation invariance is vital for generative models
on graphs'' \cite{liao2019efficient}.
GRAN is a recent efficient auto-regressive model ``generating large graphs of up to 5k vertices'' by generating a sequence of vectors representing ``one row or one block of rows'' of the lower triangular matrix at a time \cite{liao2019efficient}.

Besides those three works, there is also structure2vec \cite{dai2016discriminative}, Graphite \cite{grover2018graphite} and GraphGAN \cite{wang2018graphgan} for generically learning models of graphs.
There exist various generative models which are mostly tailored specifically to applications such as molecular designs.
We found \cite{bjerrum2017molecular} (RNN), \cite{gupta2018generative} (RNN), \cite{jin2018junction} (VAE), \cite{bjerrum2018improving} (VAE), \cite{you2018graph} (PI), \cite{li2019deepchemstable} worthy to mention here.

We also want to refer to two frameworks which provide a lot of tools for differentiable computation on graph structures such as the Deep Graph Library \cite{wang2019dgl}, Euler \cite{alibaba2019} and StellarGraph \cite{stellargraph2018}.


\section{Graph Construction Sequences}\label{sec:construction-sequences}
To learn the hidden distribution of graphs given a set of exemplary graphs we need a suitable representation from which we can learn from.
Using the flattened adjacency matrix of the graph would be one approach but ``suffers from serious drawbacks:
it cannot naturally generalize to graphs of varying size, and requires training on all possible vertex permutations or specifying a canonical permutation'' \cite{you2018graphrnn}.

We follow a similar approach as in \cite{you2018graphrnn} and \cite{li2018learning} and learn a distribution of graphs from a sequence we call \textit{actionable construction sequence}\footnote{Like with all representations of graphs there is no \textit{easy} canonical representation of a graph, otherwise the graph isomorphism problem could be solved in polynomial time.}.
In other contexts of network science this representation is also viewed under the perspective of graph evolution.


\begin{figure*}
  \begin{minipage}[c]{0.47\linewidth}
    \includegraphics[width=\linewidth]{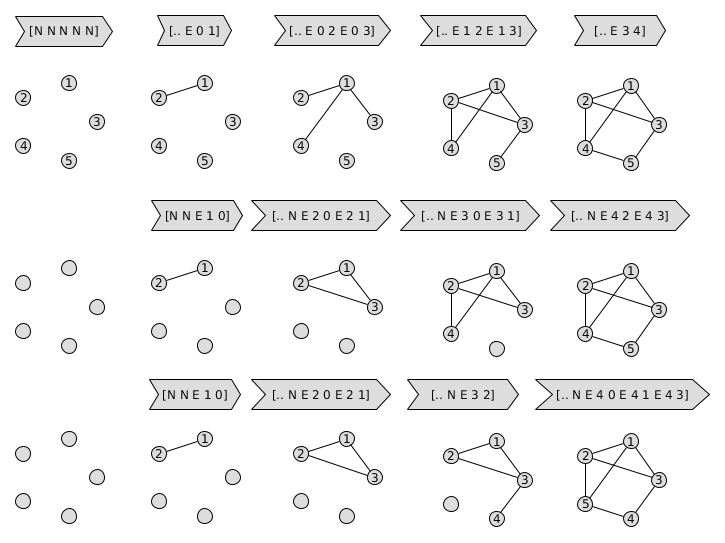}
  \end{minipage}\hfill
  \begin{minipage}[c]{0.52\linewidth}
    \caption{
		Variants for obtaining construction sequences representing the same graph (under ismorphism).
		The first graph evolution shows the steps of an Erdős-Rényi model which at first generates five vertices and then adds edges in order of the vertex numbers with an unknown probability $p$.
		The second graph evolution shows a traversal with breadth-first-search (\textit{bfs}) through the same graph and the third graph evolution shows a traversal with depth-first-search (\textit{dfs}).
		Notice, that \textit{dfs} comes up with a different vertex ordering than \textit{bfs} at the end (vertices four and five are flipped).
    } \label{fig:construction-sequences}
  \end{minipage}
\end{figure*}

An action denotes a graph operation on an initially empty graph.
Each operation can be parameterized by a fixed number of values in the sequence and thus has at least length one but at most a finite number of values.
Each action modifies the current graph by applying the graph operation on it.
Many probabilistic graph models can be represented with two graph operations:
adding a node and adding an edge.
Models such as the Watts-Strogatz model additionally require the operation to remove nodes.
All probabilistic graph models examined in our work can be learned by those three operations of adding a node, adding an edge or removing a node.
The construction sequence of a graph can be seen as the transition sequence of a (finite) state machine, but with parameterized decisions in each state -- making it a state machine with memory.

To describe a graph construction sequence we label those operations \textbf{N} (add \textbf{n}ode), \textbf{E} (add \textbf{e}dge) and \textbf{R} (\textbf{r}emove node).
Technically the labels are encoded as natural numbers in the construction sequence along with all other information which are also natural numbers.
One can easily extend the model to also learn to remove edges or perform other graph operations but it has to be considered that this adds more state decisions to the model and thus more learnable parameters.
While \textit{N} needs no further parameters, \textit{E} is followed by two parameters indicating the source and target vertex and \textit{R} is followed by one parameter denoting the affected vertex which will be removed from the graph.
Those three operations are sufficient for the most well-known probabilistic models of graphs such as the ones of Erdős-Rényi, Barabasi-Albert or Watts-Strogatz.

The null or empty graph is represented by the empty sequence $\left[~\right]$.
With the introduced labels, a graph with a single node is represented by $\left[ N \right]$ and a graph with two connected nodes with $\left[ N ~ N ~ E ~ 0 ~ 1 \right]$.
The representation encodes undirected graphs if not otherwise stated.
For a directed graph with two edges between two nodes the sequence extends to $\left[ N ~ N ~ E ~ 0 ~ 1 ~ E ~ 1 ~ 0 \right]$.

We consider three variants through which we obtain construction sequences of graphs: firstly, through the \textbf{construction process} of the particular probabilistic model, secondly through a breadth-first-search \textbf{traversal} of any given graph, and thirdly through a depth-first-search traversal of any given graph.


\paragraph{Construction Process}
Given a probabilistic model for graphs -- such as the Erdős-Rényi model with a finite number of vertices -- one can follow the algorithmic rules to build a construction sequence for each draw from the model.
If the model starts with an initial number of $n_{v_0}$ vertices the construction sequence starts with $\left[ N \right] * n_{v_0}$ -- which we understand as repeating the sequence of contained operations repeated $n_{v_0}$ times.
In the appendix we provide peusodcode for generating construction sequences following the rules of the probabilistic model of Barabasi-Albert.

\paragraph{Graph Traversal}
In application cases we usually do not know the underlying process used to come up with the observed graphs.
Thus, we need to transform graphs into construction sequences and we can do this by using common graph traversal algorithms.
The nature of the chosen traversal algorithm has most likely a major impact on our subsequent deep graph generator model (DeepGG) as the decision functions for state transitions in DeepGG are learned based on exactly those exemplary steps.

\section{Deep Graph Generator}\label{sec:deepgg}
We follow the sequential and auto-regressive generation model ``Deep Generative Models of Graphs'' (DGMG) of Li et al. \cite{li2018learning} to learn decisions from construction sequences based on a representation of the current graph state.
Our generative model -- which we will refer to as \textit{DeepGG} -- has three major differences to DGMG:
(1) we use a more relaxed notion of a finite state machine in which the model can learn to add edges from a source vertex at any step and not just at the point in which the source vertex was added to the graph, and
(2) to avoid exploding losses we reduce the negative log-likelihood loss from choosing a source or target vertex in a growing graph by dividing it with the logarithmic number of vertices at that moment, and
(3) we describe the possibility to add another state to also learn to remove vertices from the graph, which is a necessary step to learn probabilistic graph models such as the Watts-Strogatz model.

The essence of learning categorical decisions for the state transitions lies in the representational power of its memory.
Memory refers to a set of real-valued vectors (embeddings) associated with a graph.
During inference steps, a graph and associated embeddings are initialized and updated.
We use the message passing framework to learn embeddings of vertices and edges to combine them to an graph embedding.
This graph embedding and possibly a context embedding of a vertex are used to learn a transition function for states.

For a graph $G = (V, E)$ with vertices $v\in V$ and edges $e\in E = \{(s,t) \in V\times V ~|~ s \text{ has edge to } t\}$ we have embeddings $\boldsymbol{h}_v\in\mathbb{R}^{H_v}$ and $\boldsymbol{h}_e\in\mathbb{R}^{H_e}$ for hyperparameters $H_v, H_e \in\mathbb{N}^+$ for which we chose $H_v = H_e = 16$ in accordance with \cite{li2018learning}.
The number of vertices is given as $n_v = |V|$.
The graph embedding is denoted as $\boldsymbol{h}_g\in\mathbb{R}^{H_g}$ with $H_g\in\mathbb{N}^+$ for which we chose $H_g = 2\cdot H_v$.
For an empty graph we set $\boldsymbol{h}_g = (0,\dots,0)$ (a zero vector).
Like every construction sequence as depicted in \autoref{fig:construction-sequences} \textit{DeepGG} starts with an empty graph and iteratively adds edges or vertices.

\begin{figure*}
  \begin{minipage}[c]{0.55\linewidth}
    \caption{
		\textit{DeepGG} is a sequential generative model based on the notion of a state machine with memory.
		As we are generating a graph step by step, the memory in our case consists of a graph structure with node and graph embeddings on which message passing updates are performed.
		The figure shows two states (dark circles) \textit{add\_node} and \textit{add\_edge} with possible state transitions prefixed with ``$a$'' between those states.
		The lighter circles depict sub-states which can read from memory to derive a categorical decision and contribute loss to the overall end-to-end model.
		Operations (or actions) from the construction sequence control which decision a state has to take given the current memory.
	}
	\label{fig:deepgg-two-states}
  \end{minipage}\hfill
  \begin{minipage}[c]{0.44\linewidth}
    \includegraphics[width=\linewidth]{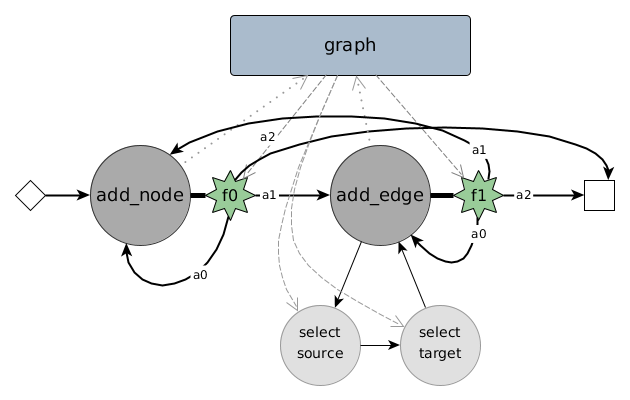}
  \end{minipage}
\end{figure*}

\paragraph{Initializing Vertices and Edges} Like in DGMG \cite[4.1]{li2018learning} we use a simple differentiable model such as a Multi-Layer Perceptron (MLP) to learn one initialization function for vertices and one for edges.
Each time \textit{DeepGG} adds a vertex $v_{new}$ or an edge $e_{new}$ an embedding for it is initialized.
While the empty graph is initialized with a zero vector, embeddings for $v_{new}$ or $e_{new}$ can be initialized from a learnable function based on the current memory state -- the graph embedding $\boldsymbol{h}_g$.
We call this initialization functions $f_{init,v}$ and $f_{init,e}$, respectively:
\[
	f_{init,v}: \mathbb{R}^{H_g} \rightarrow \mathbb{R}^{H_v}
\]
In case of a MLP with non-linear activation function $\sigma$ this could be a function mapping a graph representation $\boldsymbol{h}_g\in\mathbb{R}^{H_g}$ to $\sigma\left(W_{init,v}\cdot \boldsymbol{h}_g + b_{init,v}\right)$.
Equivalently we define	$f_{init,e}: (\mathbb{R}^{H_v}, \mathbb{R}^{H_v}) \rightarrow \mathbb{R}^{H_e}$ with $(\boldsymbol{h}_s, \boldsymbol{h}_t)\in\mathbb{R}^{H_v}\times\mathbb{R}^{H_v}\mapsto \sigma\left(W_{init,e}\cdot (\boldsymbol{h}_s \oplus \boldsymbol{h}_t) + b_{init,v}\right)$.
The learnable models $f_{init,v}$ and $f_{init,e}$ are often called sub-modules.
Parameters of $f_{init,v}$ and $f_{init,e}$ such as $W_{init,v}$, $b_{init,v}$, $W_{init,e}$ and $b_{init,e}$ are then learnable parameters of \textit{DeepGG} and their dimensions can be easily inferred by definition of our parameter choices.

\paragraph{Deep State Decisions} We identify a state with index $\kappa\in\mathbb{N}$ as $s_{\kappa}$.
Its transition decision function draws from a categorical distribution based on an embedding vector which is either a graph embedding $\boldsymbol{h}_g$, a composition of vertex embeddings $\boldsymbol{h}_v$ or a mixture of both.
Therefore, its input dimension is denoted with $H_{s_{\kappa}}$ and for each state it has to be described how to aggregate information from the memory to the particular input embedding from which the decision is derived.
The decision function $f_{s_{\kappa}}$ is given as
\[
	f_{s_{\kappa}}: \mathbb{R}^{H_{s_{\kappa}}} \rightarrow \{1,\dots,n_{s_{\kappa}}\}
\]
in which $n_{s_{\kappa}}$ is the number of possible categorical actions for that state.
During learning, the cross-entropy loss of this state decision and the observed action from the construction sequence contributes to the overall optimization objective.

A critical part of sequential state models such as \textit{DGMG} and \textit{DeepGG} besides their state transition decisions are additional objectives to learn making a choice from a growing space -- such as choosing a target vertex from a growing graph.
While \textit{DGMG} makes only target vertex decisions from the recently added vertex, \textit{DeepGG} makes two decisions when adding an edge to the graph:
one for a source and one for a target vertex.
This makes the model more generic and facilitates to more precisely follow a possibly underlying construction rule.
For example, in the Erdős-Rényi model first $n_v$ vertices are created and then edges are ``activated'' based on a probability and in order of their enumeration.
\textit{DGMG} would have arguably struggles learning from construction sequences following this process but most likely could learn from this distribution of graphs when traversing them with \textit{bfs} or \textit{dfs}.
On the other hand, more decisions with growing choices also increases the number of learnable parameters and adds an additional objective.
We compensate for exploding losses which we observed in a first version of \textit{DeepGG} by dividing the cross-entropy loss with the logarithm of possible vertex choices -- which will arguably also penalize early wrong choices over wrong choices made later on a large graph.
For one vertex choice $\delta\in\mathbb{N}$ as $c_{\delta}$ we have a categorical choice based on the current graph state given as:
\[
	f_{c_{\delta}}: \mathbb{R}^{H_{c_{\delta}}} \rightarrow \{1,\dots,n_v\}
\]
Note, that with a growing graph as memory during the generation process $n_v$ grows linearly.
When adding edges, we used two vertex choice functions $f_{c_1}$ and $f_{c_2}$ which can be seen in \autoref{fig:deepgg-two-states} as \textit{select source} and \textit{select target}.
For the memory readout we chose $H_{c_{\delta}} = H_g + H_v + H_v$ for each vertex $v$ under consideration and used the current graph embedding $\boldsymbol{h}_g$, the vertex embedding $\boldsymbol{h}_v$ and an additional contextual vertex $v_c$ embedding $\boldsymbol{h}_{v_c}$.

Equivalently to the initialisation functions for vertices and edges above, we chose single-layered MLPs as differentiable and learnable functions for all $f_{s_{\kappa}}$ and $f_{c_{\delta}}$.

\paragraph{Memory Update} In each step of a construction sequence \textit{DeepGG} and \textit{DGMG} compute a graph embedding based on the current vertex embeddings in memory.
While \textit{DGMG} aggregate a graph embedding with a ``gated sum $R(\boldsymbol{h_v}, G)$'' \textit{DeepGG} uses a gated vertex embedding in conjunction with a graph convolution \cite{kipf2016semi} which is then reduced to its mean.
For the gated vertex embedding we use a sub-module $f_{reduc}$:
\[
	f_{reduc}: \mathbb{R}^{H_v} \rightarrow \mathbb{R}^{H_{r}}, \boldsymbol{h}_v \mapsto \sigma\left( W_{reduc}\cdot\boldsymbol{h}_v + b_{reduc} \right)
\]
for which we chose $H_{r} = 7$ which then specifies the feature dimensions of the subsequent graph convolution.
We chose the mean instead of the sum as in \textit{DGMG} because we observed exploding sums for resulting graph embeddings.
Given the adjacency matrix $A\in\mathbb{R}^{n_v\times n_v}$ of a graph and input features of dimension $\mathbb{R}^{d_{in}}$ a graph convolution computes $f_{conv}: \mathbb{R}^{n_v\times d_{in}} \rightarrow \mathbb{R}^{n_v\times d_{out}}$.
\[
	f_{conv}(H^{(l)}, A) = \sigma\left( D^{-\frac{1}{2}} \hat{A} D^{-\frac{1}{2}} H^{(l)}W^{(l)} \right)
\]
with $\hat{A} = A + I$ (see \cite{kipf2016semi} for details).
We used one single graph convolution $f_{conv}$ with reduced vertex representations for each vertex in graph $G$, making up $H^{(l)}$.
$A$ is the graphs adjacency matrix.

When the current graph $G$ (the memory) is modified, we perform an update step based on the message passing framework \cite{gilmer2017neural} to update vertex representations.
The graph propagation (update) step in \textit{DeepGG} is kept equivalent to \textit{DGMG}.
Two propagation rounds $\nu_{rounds}$ of gated recurrent units are used and vertex embeddings are updated as illustrated in \cite[figure 2]{li2018learning}.

\paragraph{Training Objective} The model draws from categorical distributions to decide on state transitions or choosing vertices as source or target for an edge.
We use a cross-entropy loss on $f_{s_{\kappa}}$ and $f_{c_{\delta}}$ and minimize the joint negative log-likelihood with stochastic gradient descent.
The learning rate is $\eta = 0.0001$, we use $\nu_{epochs} = 8$ number of epochs, $\nu_{rounds} = 2$ rounds of propagations and $\nu_{n_{v\_max}} = 150$ as a hard threshold for generating a maximum number of vertices and stopping the generative process.
During generative inference a minimum value $\nu_{n_{v\_min}}$ can be specified to restrain the state machine from transitioning to a halting state before reaching a lower bound in the number of vertices.

\section{Experiments}\label{sec:experiments}
Learning distributions of graphs is not only difficult because of the complexity of the combinatorial nature of graphs but also difficult to assess properly.
Notably, due to the graph isomorphism problem we possibly require a computational expensive effort to compare two sets of graphs for overlapping isomorphic graphs.
Although our interest is not to obtain isomorphic graphs, we are faced with this or related complex problems when formulating a notion of equivalence classes of graphs.
Because we are interested in reproducing certain properties of a hidden distribution of graphs we need to investigate on if and which of those properties can statistically significantly be reproduced.

We computed 69 total \textit{DeepGG} instances of which 24 instances have been trained on datasets with bfs-traversal, 39 with dfs-traversal and six based on the construction process of the underlying probabilistic model.
24 instances have been computed with datasets based on the Erdős-Rényi model, 26 instances on the Barabasi-Albert model and 19 on the Watts-Strogatz model.
Across all computations we used the same objective, the same learning rate, the same number of epochs and other hyperparameters as stated in \autoref{sec:deepgg}.
For each computation instance we sampled new sets of graphs of size $n_v = 50$ from the ER-model with $p = 0.2$, from the WS-model with $k = 10, p = 0.2$ and from the BA-model with $m = 3$.

We argue, that it is important to investigate on multiple levels of distributions of graph properties to constitue that a certain hidden process could be reflected.
In network science, complex networks are compared based on the degree distribution, clustering coefficient \cite{watts1998collective} or on the average path length distribution (or similar aggregated statistics of the underlying properties).

\begin{figure*}
  \begin{minipage}[c]{0.49\linewidth}
	\centering
    \includegraphics[width=\linewidth]{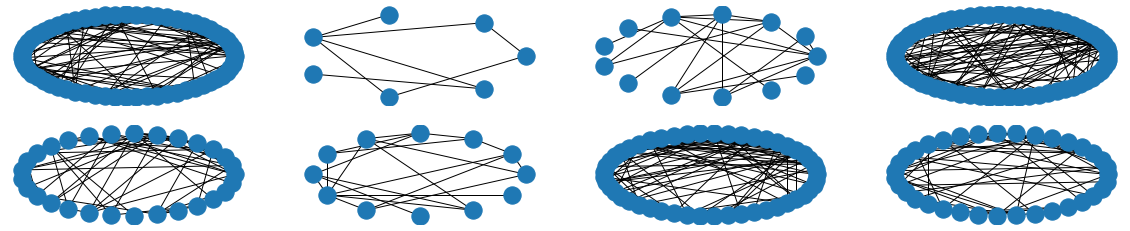}\\
	{\color{gray}\small\#t0175785 [BA-model] [process]}\protect\\
    \includegraphics[width=\linewidth]{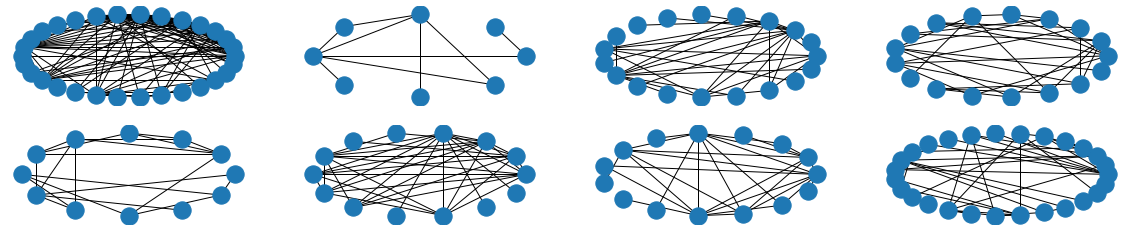}\\
	{\color{gray}\small\#t0408015 [BA-model] [dfs]}
  \end{minipage}\hfill
  \begin{minipage}[c]{0.49\linewidth}
	\centering
    \includegraphics[width=\linewidth]{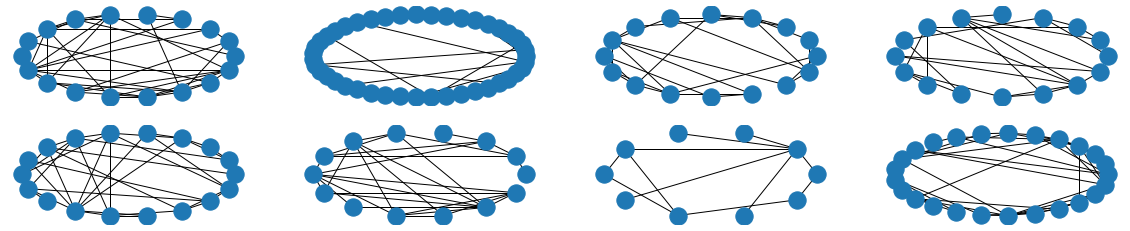}\\
	{\color{gray}\small\#t0175786 [WS-model] [dfs]}\protect\\
    \includegraphics[width=\linewidth]{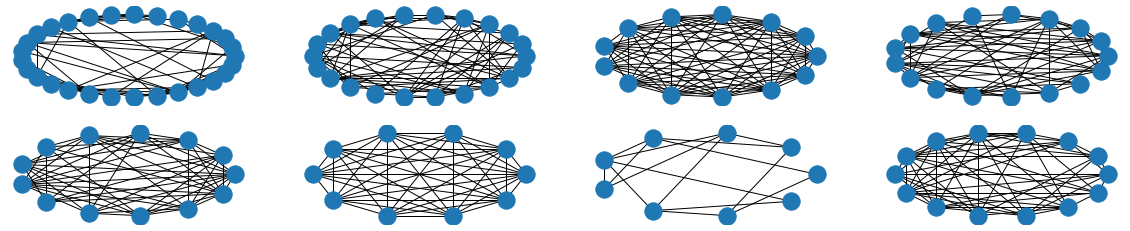}\\
	{\color{gray}\small\#t0574307 [ER-model] [dfs]}\protect\\
  \end{minipage}
  \caption{
    Exemplary samples from computed \textit{DeepGG} instances.
	The id is a shortened computation timestamp.
	Layouts are circular to provide a visual impression of the density.
	The original distribution is based on samples from either an Erdős-Rényi, a Watts-Strogatz or a Barabasi-Albert probabilistic model.
	Used traversal method are breadth-first-search (bfs), depth-first-search (dfs) or based on the process of the model as described in \autoref{sec:construction-sequences}.
  } \label{fig:samples}
\end{figure*}

\paragraph{Degree Distribution} The three considered probabilistic graph models have distinct degree distributions.
Barabasi-Albert is known to have a scale-free degree distribution.
Erdős-Rényi has a binomial distribution $p_k \sim B(n_v, ER_p)$ which in our case gives $E(p_k) = 10$.
The degree distribution of Watts-Strogatz graphs are similar to random graphs in shape.
For selected \textit{DeepGG} instances we find that they are able to reproduce the degree distribution very closely.
However, we also encounter instances which follow a scale-free distribution when trained on ER-models.
Similar to \textit{DGMG} we find a close relationship to the degree distribution of scale-free networks but do not observe the same behaviour for WS-graphs.


\paragraph{Average Path Length Distribution}
We observe a mean of $1.75$ for the distribution of average path lengths of \textit{DeepGG} learned on ER-graphs.
This observed mean of generated graphs indeed gets close to the mean in the dataset of $1.91$.
According to the analytic form given by \cite{fronczak2004average} it must be $l_{ER} = \frac{ln(n_v)-\gamma}{ln(p\cdot n_v)}+\frac{1}{2}$ with $\gamma$ being the Euler-Mascheroni constant.
We have $l_{ER} = \frac{ln(50)-\gamma}{ln(0.2\cdot 50)}+\frac{1}{2} \approx 1.95$.
For Barabasi-Albert graphs the analytical average path length is $l_{BA} = \frac{ln(n_v)-ln(m/2)-1-\gamma}{ln(ln(N))+ln(m/2)}+\frac{3}{2}$ for which we get $l_{BA} \approx 2.59$.
In our dataset of BA-graphs we observe an averaged average shortest path length of $2.29$ while \textit{DeepGG} resembles $2.61$, again with a high standard deviation of $1.50$.


\begin{figure*}
  \begin{minipage}[c]{0.48\linewidth}
    \includegraphics[width=\linewidth]{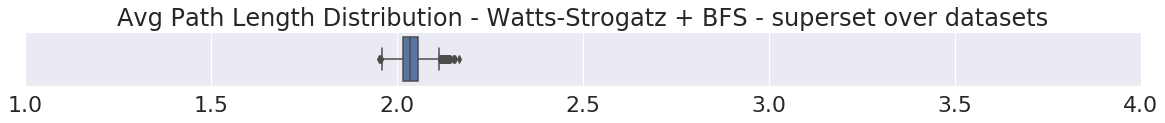}
    \includegraphics[width=\linewidth]{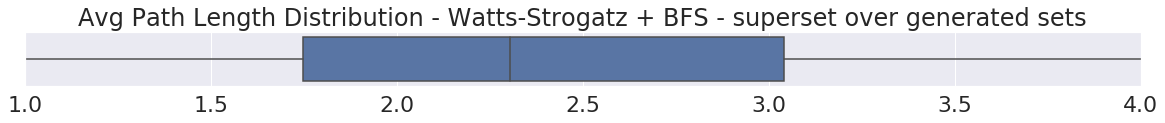}
  \end{minipage}\hfill
  \begin{minipage}[c]{0.48\linewidth}
    \includegraphics[width=\linewidth]{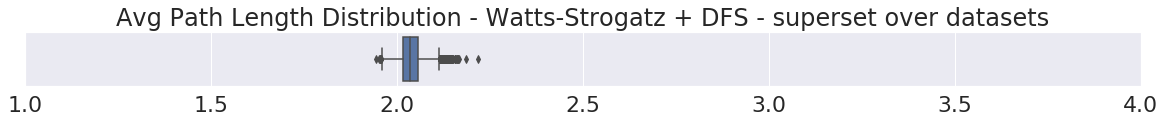}
    \includegraphics[width=\linewidth]{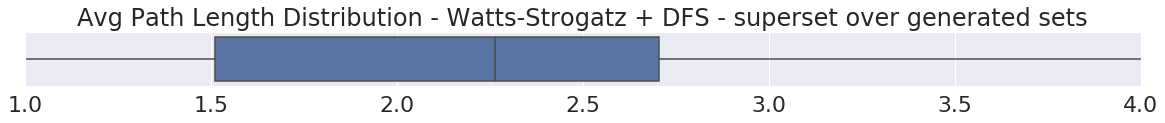}
  \end{minipage}
    \caption{
      For multiple instances of \textit{DeepGG} trained on Watts-Strogatz graphs we computed the average path length of each of the 1,000 graphs used for training for both bfs- and dfs-traversal.
      Equivalently, we computed the average path length of each generated graph from the model (on average 171 for the BFS-instances and 170 for the DFS-instances).
      Seven \textit{DeepGG} instances have been trained on bfs-traversals, twelve on dfs-traversals.
      Watts-Strogatz+BFS constitute a superset in which we generated 1,199 graphs, Watts-Strogatz+DFS a superset in which we computed 2,034 graphs.
      As to no surprise the average path length distributions for both supersets over the datasets are visually equivalent.
      However, we have to ascertain that the average path length distributions for generated graphs can not only be matched its dataset but also have to point out that the resulting superset distribution for different traversal choices for construction sequences differs significantly.
    }
    \label{fig:avg-path-length-supersets}
\end{figure*}

\paragraph{Computation Times} Overall the computation time over training datasets of 1,000 graph samples with each large construction sequences has been very high.
We used GPU acceleration but did not perform batching.
The average computation time was roughly 60 hours with a minimum of 20 hours and a maximum of 120 hours.
Instances trained on BA-graphs took only 40 hours on average with a low standard deviation while WS-graphs took 65 hours and ER-graphs over 70 hours on average.
We also observed significantly different distribution between computation times when using dfs- or bfs-traversal.

\subsection{Findings in our Experiments}
First of all, we do find \textit{DeepGG} instances which generate new graphs with property distributions close to the underlying probabilistic models.
We even manage to generate larger graphs than reported by Li et al. of up to a few hundred vertices, although the recently published \textit{GRAN} reportedly achieves to generate several thousand of vertices.
Looking at single instances visually one could easily think that \textit{DeepGG} (or \textit{DGMG}) instances learn meaningful distributions of graphs.
Repeated experiments and reasoning from the nature of both \textit{DeepGG} and \textit{DGMG}, however, leads us to the conjecture that the models are biased towards scale-free random graphs and can not repeatedly resemble the behaviour of the targeted probabilistic model -- if not supervised carefully.

The choice of representation of graphs has a significant impact on learning distributions over graphs as already stated in \cite{you2018graphrnn} or \cite{liao2019efficient}.
In case of our chosen representation with construction sequences we have a huge number of possible permutations over a random ordering of vertices for a single graph.
We investigated on three variants of canonical traversal methods over random orderings of vertices and observed different resulting distributions for average shortest path lengths and vertex degrees.
For depth-first-search we observe a higher variance in computation times than for breadth-first-search.
The difference in the mean of their computation times could be a result to the low number of overall computed instances over a rather high number of hyperparameter choices.
We further observe a high computation time for ER-models over WS-models and BA-models.

Especially the message passing component to obtain vertex and graph embeddings is a computational expensive part of the model and we see possibilities for improvement there.
The growing number of vertices further leads to a growing categorical choice space for $f_{c_{\delta}}$ which we identify as an essential step towards learning other degree distributions than the distribution of a BA-model.
We see a possibility of ``almost random'' transitions through the state machine which might result in the generation of scale-free random networks and we will investigate further in this decision process.


\section{Conclusion}\label{sec:conclusion}
We described a generative model for learning distributions of graph as a generalization of the model given in \cite{li2018learning}.
The generalization made it possible to learn distributions of graphs based on the graph construction process of the \textit{Watts-Strogatz} model and we are confident that \textit{DeepGG} reduced bias towards scale-free graphs.
Our experiments showed, that often an underlying construction process could not be learned although single instances showed promising resuls -- visually and based on graph property distributions.
We emphasize that generative models of graphs need to be evaluated on multiple levels to reach statistical significant evidence and focussed on providing such experimental setups.

\textit{DeepGG} was rendered under the perspective of a \textit{deep state machine}, which can be seen as a state machine with learnable transitions and a memory.
The decision process of such a state machine can be analysed more transparently as compared to other common end-to-end models.

We hope to contribute valuable insights for learning distributions of graphs and expect to (1) further investigate generative models of graphs and (2) apply the recently developed models in promising domains.

\section*{Broader Impact}

This work investigates on learning distributions of graphs from a set of exemplary graphs with the aim to generate new graphs from the same or closely similar hidden distribution.
Such graph generators have use cases in molecular design, network analysis, program synthesis and probably many more fields which is why we move in the direction of focussing on evaluating generic methods for learning distributions of graphs.
Beneficiaries can be considered everyone who is interested in automating design of complex underlying graph structures.
Of course, such methods could then also be used for drug discovery or other designs which at the wrong end can have negative ethical aspects.
We argue, however, that this is a conflict which has to be solved on another societal level and is a result of a more general drive towards automation.
As the described problem on graph distributions is mostly of theoretical nature, we do not see that the method leverages bias of ethical concern.
In general, research on learning distributions of graphs will improve automation and design in various domains.
We see mostly societal benefits from this automation process.


\begin{ack}
We thank Jörg Schlötterer for valuable discussions on our work.

\end{ack}

\bibliographystyle{ACM-Reference-Format}
\bibliography{content/bibliography}

\newpage
\section*{Appendix}
\begin{algorithm}
\SetKwInOut{Input}{Input}
\SetKwInOut{Output}{Output}
\SetKwData{S}

\underline{function BA-seq} $(n, m)$\;
\Input{Two nonnegative integers $n$ and $m$}
\Output{Construction sequence for $BA(n, m)$}
\BlankLine
$seq \leftarrow \left[N\right] * m$\;

$t \leftarrow \left[0,\dots,m-1\right]$\;
$r \leftarrow \left[\right]$\;
$cs \leftarrow m$\;
\While{cs < n}{
	$seq \leftarrow seq + \left[ N \right]$\;
    \lForEach{$ct$ in $t$}{
        $seq \leftarrow seq + \left[E ~ cs ~ ct\right]$
	}

	$r \leftarrow r + t$\;
	$r \leftarrow r + \left[cs\right] * m$\;
	$t \leftarrow random\_subset(r, m)$\;

    $cs \leftarrow cs + 1$\;
}
\KwRet{$seq$}
\caption{Example for generating a construction sequence based on the Barabasi-Albert probabilistic graph generator.}
\label{algo:barabasi-albert-conseq}
\end{algorithm}

\begin{figure*}
  \begin{minipage}[c]{0.48\linewidth}
    \includegraphics[width=\linewidth]{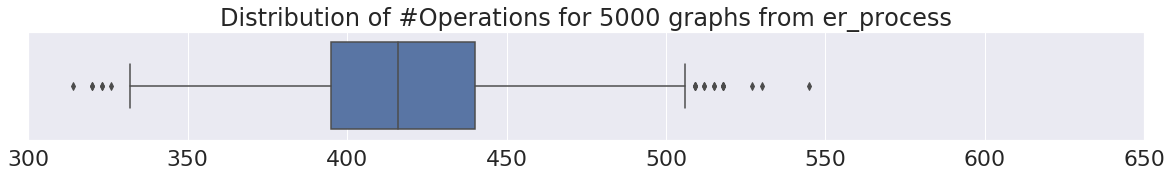}
    \includegraphics[width=\linewidth]{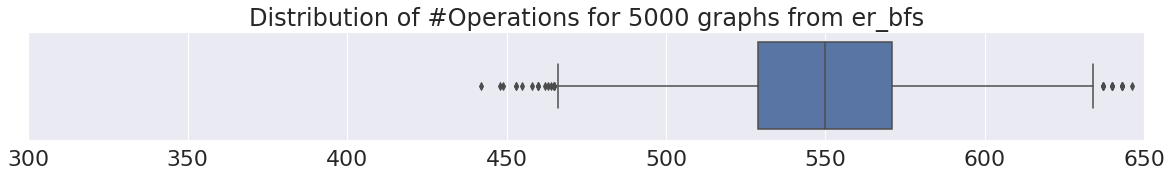}
    \includegraphics[width=\linewidth]{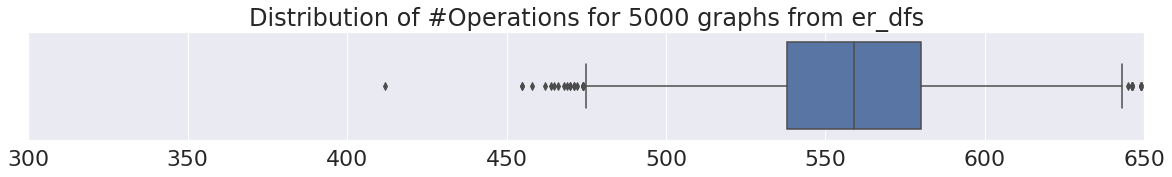}
  \end{minipage}\hfill
  \begin{minipage}[c]{0.48\linewidth}
    \includegraphics[width=\linewidth]{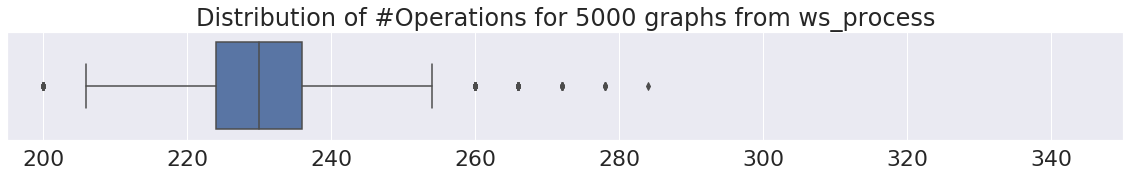}
    \includegraphics[width=\linewidth]{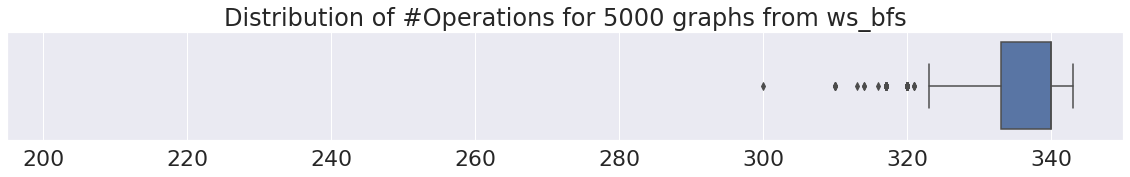}
    \includegraphics[width=\linewidth]{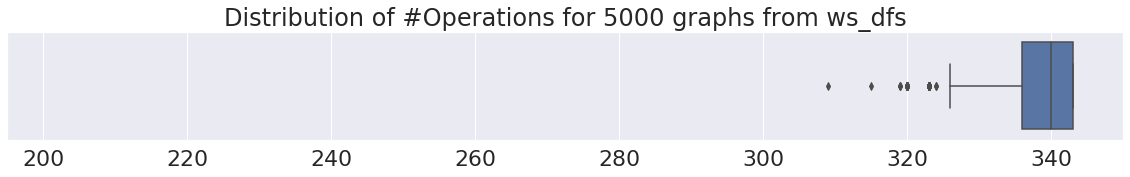}
  \end{minipage}
  \caption{
    Distributions of construction sequence lengths for two probabilistic models Erdős-Rényi (left side) and Watts-Strogatz (right side) for the three mentioned variants ``construction \textit{process}'', ``\textit{\textbf{d}epth-\textbf{f}irst \textbf{s}earch}'' and ``\textit{\textbf{b}readth-\textbf{f}irst \textbf{s}earch}''.
  } \label{fig:num-operations}
\end{figure*}

\end{document}